\documentclass[]{config/antgroup}
\usepackage{config/antgroup}

\newcommand*\justify{%
  \fontdimen2\font=0.4em
  \fontdimen3\font=0.2em
  \fontdimen4\font=0.1em
  \fontdimen7\font=0.1em
  \hyphenchar\font=`\-
}

\renewcommand{\texttt}[1]{%
  \begingroup
  \ttfamily
  \begingroup\lccode`~=`/\lowercase{\endgroup\def~}{/\discretionary{}{}{}}%
  \begingroup\lccode`~=`[\lowercase{\endgroup\def~}{[\discretionary{}{}{}}%
      \begingroup\lccode`~=`.\lowercase{\endgroup\def~}{.\discretionary{}{}{}}%
      \catcode`/=\active\catcode`[=\active\catcode`.=\active
        \justify\scantokens{#1\noexpand}%
        \endgroup
      }

      \PassOptionsToPackage{numbers, compress}{natbib}

      \usepackage{graphicx}
      \usepackage{todonotes}
      \usepackage{multirow}
      \usepackage{hyperref}
      \usepackage[capitalize,noabbrev,nameinlink,sort]{cleveref}
      \usepackage{subcaption}
      \usepackage{multicol}
      \usepackage{array}
      \usepackage{enumitem}
      \usepackage{algorithm}
      \usepackage{algpseudocode}
      \usepackage{tabularx}
      \usepackage{makecell}
      \usepackage{xspace}
      \usepackage{colortbl}
      \usepackage{fontawesome5}
      \usepackage{pifont}
      \usepackage[normalem]{ulem}
      \useunder{\uline}{\ul}{}
      \usepackage{tcolorbox}
      \usepackage{tablefootnote}
      \usepackage{booktabs,tabularx,threeparttable,array}
      \usepackage[table]{xcolor}
      \usepackage{arydshln}

      \usepackage{amsmath}
      \usepackage{amssymb}
      \usepackage{amsfonts}                               
      \usepackage{amsthm}
      \usepackage[mathcal]{eucal}
      \usepackage{mathrsfs}
      \usepackage{bm}                                     
      \usepackage{blkarray}                               
      \usepackage{nicefrac}                               
      \usepackage{bbm}

      \usepackage{wrapfig}
      \usepackage{graphicx}                               
      \usepackage{caption}
      \usepackage{cleveref}

      \usepackage{tikz}
      \usepackage{circuitikz}
      \usetikzlibrary{patterns}
      \usetikzlibrary{positioning,calc,fit,decorations.pathmorphing,shapes.geometric, shapes.gates.logic.US, calc}
      \usetikzlibrary{arrows,arrows.meta,decorations.markings,shapes,shapes.arrows}
      \usetikzlibrary{decorations,decorations.pathreplacing}
      \usetikzlibrary{backgrounds}
      \usepackage{pgfplots}
      \usepackage{pgfplotstable}
      \usepgfplotslibrary{groupplots}
      \usepackage{scalefnt}
      \pgfplotsset{compat=newest}
      \usepackage{xcolor}

      \definecolor{firstcolor}{HTML}{C3423F}
      \definecolor{secondcolor}{HTML}{2A4B8C}
      \definecolor{aworld_blue}{HTML}{4e81ff}
      \definecolor{aworld_cyan}{HTML}{41d7fa}
      \definecolor{aworld_teal}{HTML}{5fede4}

      \hypersetup{
        colorlinks=true,
        linkbordercolor=aworld_blue,
        citebordercolor=aworld_cyan,
        urlbordercolor=aworld_teal
      }

\def\eqref#1{equation~\ref{#1}}

\def\1{\bm{1}}

\DeclareMathAlphabet{\mathsfit}{\encodingdefault}{\sfdefault}{m}{sl}
\SetMathAlphabet{\mathsfit}{bold}{\encodingdefault}{\sfdefault}{bx}{n}

\begin{document}

\title{Recon-Act: A Self-Evolving Multi-Agent Browser-Use System via Web Reconnaissance, Tool Generation, and Task Execution}

\author{
  Kaiwen He$^{1,\ast}$, 
  Zhiwei Wang$^{1,\ast}$, 
  Chenyi Zhuang$^{1}$,
  Jinjie Gu$^{1}$
}

{\renewcommand\thefootnote{}
  \footnotetext{$^\ast$Equal contributions.}
}

\affiliation{$^1$\raisebox{-0.2em}{\includegraphics[height=1.1em]{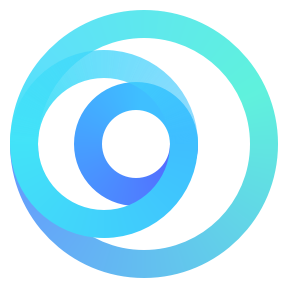}}AWorld Team, Inclusion AI\quad}
\maketitle

\begin{center}
  \vspace{-1.5em}
  \href{https://github.com/inclusionAI/AWorld/tree/main/examples/visualwebarena}{\faGithub\ \texttt{https://github.com/inclusionAI/AWorld/tree/main/examples/visualwebarena}}
  \vspace{0.5em}
\end{center}

\begin{abstract}
Recent years, multimodal models have made remarkable strides and 
pave the way for intelligent browser-use agents.
However, when solving tasks on real-world webpages 
in multi-turn, long-horizon trajectories, 
current agents still suffer from disordered action sequencing and 
excessive trial-and-error during execution. 
This paper introduces Recon‑Act, 
a self-evolving multi-agent framework grounded in
Reconnaissance–Action behavioral paradigm.
The system comprises a Reconnaissance Team and an Action Team: 
the former conducts comparative analysis and tool generation, 
while the latter handles intent decomposition, tool orchestration, and execution. 
By contrasting the erroneous trajectories with successful ones, 
the Reconnaissance Team infers remedies, and abstracts them
into a unified notion of ``generalized tools'',
either expressed as hints or as rule-based codes,
and register to the tool archive in real time.
The Action Team reinference the process empowered with these targeting tools,
thus establishing a closed-loop training pipeline of 
data–tools–action–feedback. 
Following the 6 level implementation roadmap 
proposed in this work, 
we have currently reached Level 3 
(with limited human-in-the-loop intervention). 
Leveraging generalized tools obtained through reconnaissance, 
Recon‑Act substantially improves adaptability to unseen websites and 
solvability on long-horizon tasks, 
and achieves state-of-the-art performance on the challenging VisualWebArena dataset.
\end{abstract}

\begin{figure}[H]
  \centering
  \begin{subfigure}[]{0.5\textwidth}
    \centering
    \includegraphics[width=\linewidth]{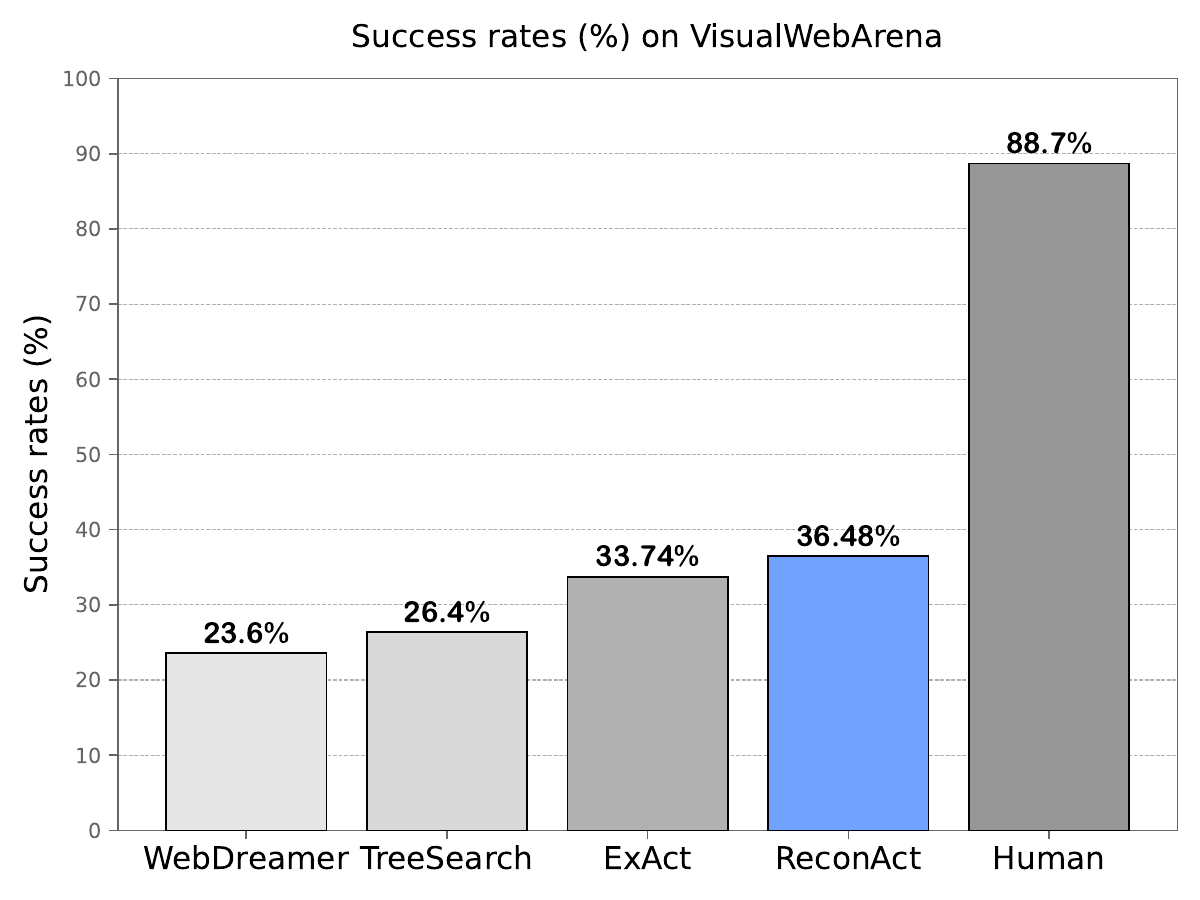}
  \end{subfigure}%
  \begin{subfigure}[]{0.5\textwidth}
    \centering
    \includegraphics[width=\linewidth]{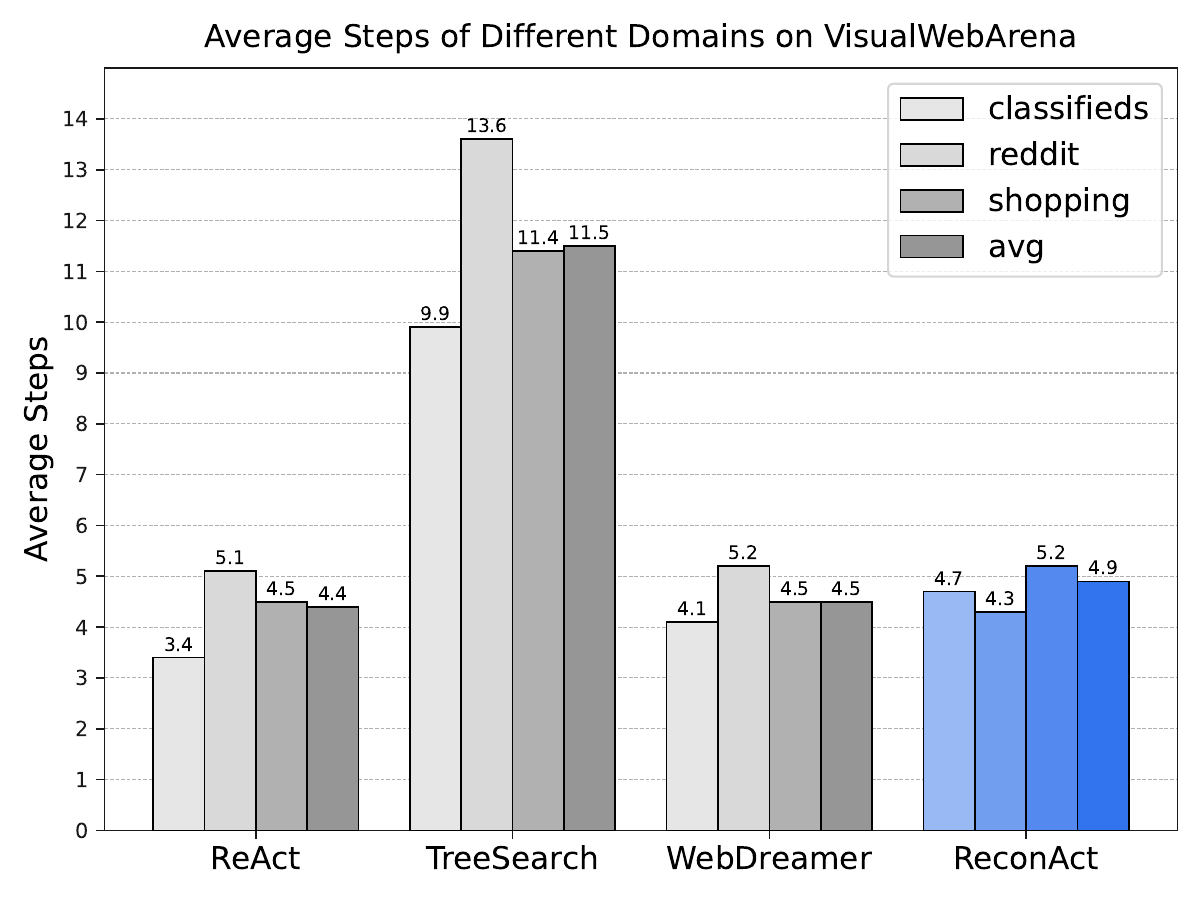}
  \end{subfigure}
  \caption{\textbf{Success Rates on VisualWebArena Dataset (Left)}
While remains a substantial gap to human performance, 
Recon-Act reaches 36.48\% success rates, outperforming the other automated agents
\citep{gu2025llmsecretlyworldmodel,koh2024treesearchlanguagemodel,yu2025exactteachingaiagents,koh2024visualwebarenaevaluatingmultimodalagents}.
  \textbf{Average Steps of Different Domains on VisualWebArena (Right)} 
Despite requiring a moderate number of steps,
Recon-Act achieves stable web navigation
with only little self‑corrective actions.
Other steps data comes from \citet{gu2025llmsecretlyworldmodel}.
}
  \label{fig:main}
\end{figure}

\section{Introduction}
\label{sec:intro}


Recently, Multimodal Large Language Models (MLLM) 
presented by 
\citet{openai2024gpt4technicalreport, qwenvl, qwen2vl, qwen2.5vl, internvl,internvl1.5,internvl2.5,wang2023cogvlm,hong2024cogvlm2}
have markedly advanced visual understanding, long‑context reasoning, 
and native tool‑use capabilities, 
laying the groundwork for autonomous browser‑use agents. 
Nevertheless, in real‑world web settings, 
multi‑turn and long‑trajectory tasks 
continue to suffer from brittle tool orchestration and 
trial‑and‑error in unfamiliar environments. 
For example, on the dataset proposed by \citet{koh2024visualwebarenaevaluatingmultimodalagents}, 
which reflects real browser‑use needs,
several state‑of‑the‑art MLLMs
still fall far short of human performance.
Recent studies have proposed methods to 
improve browser-use ability on complex tasks.
However, some GUI-based studies 
\citep{liu2025pcagenthierarchicalmultiagentcollaboration,gu2025uivenustechnicalreportbuilding,ye2025mobileagentv3fundamentalagentsgui,lian2025uiagileadvancingguiagents} 
has not yet been designed specifically for browser environment.
Dynamic planning methods
\citep{koh2024treesearchlanguagemodel,yu2025exactteachingaiagents}
can explore solution paths autonomously,
but often require a large
amount of simulation to identify optimal actions, 
resulting in long trajectoris.
\citep{yu2025exactteachingaiagents}.

Looking at the overall picture, 
\citet{wang2025surveylargelanguagemodel} 
set agent architectures as four components.
They are profile, memory, planning, and action,
and the last of which includes the capability to invoke external tools. 
Drawing on our analysis of browser-use–specific datasets 
\citep{koh2024visualwebarenaevaluatingmultimodalagents,deng2023mind2webgeneralistagentweb,zheng2024gpt4visiongeneralistwebagent},
on the one hand we find that current agents remain limited in their 
comprehension and reasoning abilities.
On the other hand, for browser-based applications in particular, 
they would benefit from external tools on 
acquiring task-relevant information or finishing some key actions.
Augmenting MLLMs’ ability to obtain information via 
external tools can alleviate 
the constraints of parametric knowledge and, 
in turn, curb hallucinations 
\citep{qu2024toollearninglargelanguage,wang2025surveylargelanguagemodel}. 
A growing body of work advances this agenda 
under the rubric of tool learning \citep{he2025gentoolenhancingtoolgeneralization,shi2025toollearningwildempowering,qin2023toolllmfacilitatinglargelanguage,schick2023toolformerlanguagemodelsteach}.
With the repid growth on coding abilities of large models
\citep{li2023starcodersourceyou,lozhkov2024starcoder2stackv2,qwen3technicalreport,hui2024qwen2,guo2024deepseekcoderlargelanguagemodel,zhu2024deepseekcoderv2}, 
We consider directly enabling the model to synthesize the tools 
it judges most appropriate and, 
on that basis, 
to formulate the most suitable solution to the problem at hand.
Considering the information density of browser environments, 
where only a subset of observations is germane to a particular task,
we design our tools to return distilled, task-salient information 
and, when appropriate, to directly yield the executable action.
Human users usually scanning the page for an overall 
picture before taking action when facing an unfamiliar web page.
Inspired by this,
we seek to extract useful information through 
a limited number of actions to guide subsequent execution.
We define this information exploration and distillation process 
as a “reconnaissance” operation, 
which involves conducting exploratory actions in the environment 
and collecting additional observational data 
when the task is not being performed properly or seems infeasible. 
Based on the insights gained, we provide recommendations to the 
task-executing agent to help it complete the task.
Such advice may take the form of hints or specialized tools 
that solve particular problems in specific contexts. 
We unify these as generalized tools.
\citet{xie2025profileawaremaneuveringdynamicmultiagent} 
establishes a multi-agent framework
consisting of a primary executor agent and 
an on‑demand guardian agent that supervises, 
validates reasoning, 
and corrects error to achieve evolution.
We propose a similar dyadic framework 
in which we substitute the ``Guardian''
to a ``Reconnaissance Team''. 
The Reconnaissance Team provides actionable guidance 
in the form of hints or dedicated tools, 
both encapsulated as generalized tools.

This paper present Recon-Act, 
a self‑evolving multi‑agent system (MAS) 
specifically designed for browser‑use tasks, 
which places tools, in a broad sense 
encompassing both rule‑based tools and tool agents, 
as the core of the iterative process. 
The positive and negative trajectories serve as sources of feedback. 
Through contrastive analysis over these instances, 
the system derives feedback signals and 
establishes a closed‑loop, data–tool–action–feedback evolutionary pipeline.
Recon-Act enables agents to acquire targeted cues and 
assistance in unfamiliar environments and 
thereby complete general tasks more effectively.

Our system can be succinctly characterized as a multi-agent system
composed of a Reconnaissance Team and an Action Team. 
Within the former,
we define two agents, that is Analyst and Coder. 
While the Action Team comprises
a Master, a Tool Manager, and an Execution Agent.
We designed a series of progressively staged hypotheses and experiments 
that incrementally operationalize components 
of this architecture across 6 levels:

\begin{enumerate}[leftmargin=12pt, nosep]
  \item \textbf{Level 1}: All components are human-operated except the Execution Agent.
  \item \textbf{Level 2}: The Master and Execution Agent are powered by a 
  vision-language model (VLM); all other components remain human-operated.
  \item \textbf{Level 3}: The Master, Execution Agent, 
  and Coder are powered by large language/vision-language models (LLM/VLM); 
  the remaining components are human-operated.
  \item \textbf{Level 4}: All components except the Analyst are powered by LLMs/VLMs.
  \item \textbf{Level 5}: All agents are powered by LLMs/VLMs.
  \item \textbf{Level 6}: An end-to-end model that can finish all the tasks.
\end{enumerate}

Because of the problem’s difficulty and current limitations in LLM/VLM reasoning, 
our implementation reaches Level 3: 
both the Analyst and Tool Manager retain a degree of human-in-the-loop intervention. 

The main contributions of this paper include:
\begin{itemize}[leftmargin=12pt, nosep]
  \item 
  We propose Recon-Act, 
  a self-evolving browser-use agent framework 
  centered on a ``reconnaissance-action'' dual-team collaboration.
  We formalize ``reconnaissance operations'' in a browser context, 
  which distill key observations from information-dense 
  web pages through a small number of exploratory actions, 
  and improves the solvability and efficiency of long-term, 
  multi-round tasks.
  \item Under Level 3 configuration, our system achieves state-of-the-art performance on the VisualWebArena dataset.
\end{itemize}

\section{Related Work}
\label{sec:related}


\subsection{GUI-Agents with browser-use ability}

Several studies leverage a more general GUI agent paradigm to solve tasks including
browser-use, GUI operation, understanding etc.
PC-Agent \citep{liu2025pcagenthierarchicalmultiagentcollaboration} 
decomposes desktop control into a three-level multi-agent 
hierarchy augmented with an Active Perception Module.
UI-Venus \citep{gu2025uivenustechnicalreportbuilding} mitigates reasoning 
drift and amplifies rare key actions via 
Self-Evolving Trajectory History Alignment and Sparse Action Enhancement, 
coupled with a data-curation pipeline that yields cleaner grounding and navigation sets.
GUI-Owl \citep{ye2025mobileagentv3fundamentalagentsgui} 
trains a single model that unifies perception, 
reasoning, and action, RL-aligned on real tasks and deployed as share-observation specialists inside Mobile-Agent-v3 for long-horizon mobile workflows. 
UI-AGILE \citep{lian2025uiagileadvancingguiagents} continues 
supervised fine-tuning (SFT) with a continuous center-reward, 
a “simple-thinking” loss, and crop-resampling to combat sparse rewards.
At test time it stitches VLM-chosen candidates from decomposed high-resolution crops.
ViGoRL \citep{sarch2025groundedreinforcementlearningvisual} 
is a reinforcement learning-based VLM that anchors each reasoning step to specific visual coordinates, producing spatially grounded traces that guide attention and, via a novel multi-turn RL framework, dynamically zooms into predicted regions for fine-grained exploration.
ICAL \citep{sarch2025vlmagentsgeneratememories} 
proposed an in-context abstraction learning framework that 
enables VLM agents to convert suboptimal demonstrations into 
high-quality training data by 
self-reflecting to derive generalized strategies and action annotations,
iteratively refined with human feedback during execution in similar environments.
Departing from the foregoing, 
we argue that in the browser environment, 
attention should not only be paid to its distinctive action space but, 
more importantly, to its environment-specific observation space, 
as doing so can substantially enhance execution performance. 
This insight led to the initial conception of observation-generation tools.

\subsection{Dynamic Planning Methods}

Recent studies adopts dynamic planning procedures.
At each decision step, the agent generates multiple candidates 
(actions, intermediate thoughts, or subplans), 
scores them using one or more evaluators 
(e.g., a value function, a reward model, or LLM-based self-evaluation or debate), 
selects the best candidate for execution, 
and iterates when necessary. 
We collectively refer to this family of approaches as 
Dynamic Planning methods, which share a similar pipeline: 
candidate generation, evaluation or scoring, 
selection or backtracking, execution.
ExAct \citep{yu2025exactteachingaiagents} 
augments MCTS with reflection reuse and multi-agent debate, 
then distills the full search loop into the model. 
\citep{koh2024treesearchlanguagemodel} performs best-first search on 
an on-the-fly interface graph scored by an LM-value function, 
yielding the first verified lift on real websites. 
Agent Q \citep{putta2024agentqadvancedreasoning} 
pairs MCTS over web pages with LM self-critique, 
turning LLM-generated step-level rankings 
into dense rewards that guide exploration without human labels.
WebDreamer \citep{gu2025llmsecretlyworldmodel} lets an LLM “dream” a 
next-state description for every candidate action, 
executes the most promising, and iterates until self-judged success. 
Across these studies, 
Recon-Act is distinguished by its tool-centric, 
reconnaissance-driven design: 
it initiates targeted exploration when progress stalls, 
distills the resulting observations into generalized 
tools (either hints or dedicated tool agents) and 
closes the loop via contrastive analysis over positive and 
negative instances to refine policies in the form of tools.
This yields a practical path to self-evolution in 
information-dense browser environments, 
complementary to advances in agentic RL.

\subsection{Agent and Tool}

Within the broader topic of agents and tools, 
progress chiefly proceeds along two directions: 
tool learning, 
namely improving a model’s ability to select and use tools, 
and tool generation 
(within the framework of this paper this corresponds to code generation).  
GenTool \citep{he2025gentoolenhancingtoolgeneralization} 
synthesizes two types of training data: 
zero-to-one generalization for queries without suitable tools,
and weak-to-strong generalization for 
queries suitable for using optimized tools.
It further proposes a two-stage fine-tuning 
procedure, optimizing tool ranking, 
then refining tool selection, to strengthen 
tool-use capability.  
AutoTools \citep{shi2025toollearningwildempowering} 
pre-encapsulates tools as callable functions and 
verifies both syntactic and runtime correctness.
At inference time, it generates code-like
invocation logic to execute tool calls and 
provides error feedback.
ToolLLM \citep{qin2023toolllmfacilitatinglargelanguage} 
constructs datasets using LLM and 
employs an automatic evaluator 
whose core algorithm is a DFS-based decision tree.
to fine-tune the model.
Toolformer \citep{schick2023toolformerlanguagemodelsteach} 
augments existing corpora to derive an API-call dataset and, 
enables the model to learn how to use external tools via fine-tuning.
On the code generation front, 
\citet{li2023starcodersourceyou,lozhkov2024starcoder2stackv2,qwen3technicalreport,hui2024qwen2,guo2024deepseekcoderlargelanguagemodel,zhu2024deepseekcoderv2}
leverage large-scale training data to train specialized coding
models across a range of model sizes.

\section{Methodology}
\label{sec:methodology}

\begin{figure}[!t]
  \centering
  \includegraphics[width=0.99\textwidth]{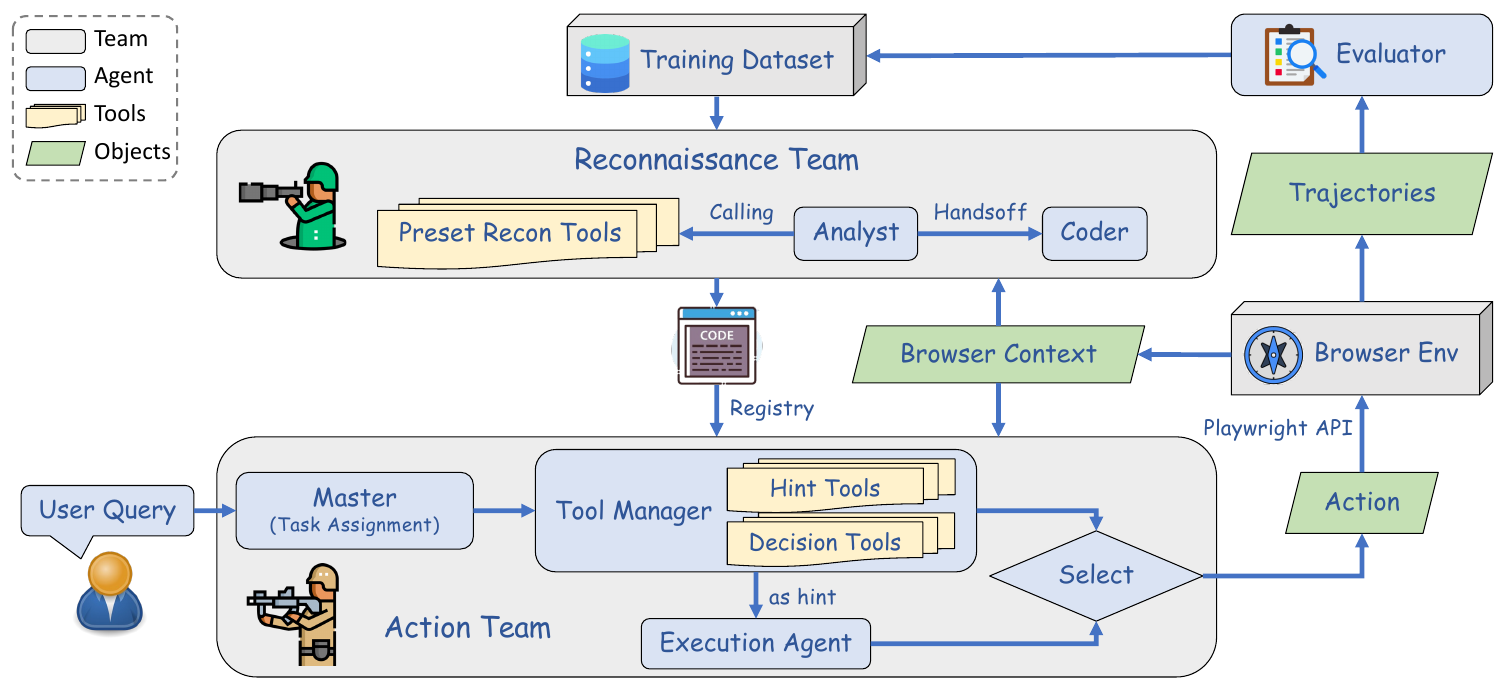}
  \caption{\textbf{System Architecture}.
  The system comprises two integrated teams: 
  Reconnaissance and Action Team. 
  Training workflow proceeds as follows. 
  A user query together with the browser context is 
  ingested by a Master Agent, 
  which invokes an appropriate agent or tool. 
  A Router then selects suitable tools or member agents to get answer.
  A Selection module consolidates the outputs into a final action, 
  which is executed in browser via the Playwright API and yields trajectory.
  An Evaluator reviews the trajectory and 
  writes the assessment back to the training set. 
  If the trajectory is still incorrect, 
  the Reconnaissance Team employs preconfigured reconnaissance tools 
  to gather additional information. 
  Its Analyst devises a plan and the Coder implements a new tool. 
  The new tool will be registried and deployed online to 
  the Action Team’s tool manager, 
  after which subsequent tasks proceed using the augmented toolset. 
  In this way, trajectories, evaluation, and training form a closed-loop, 
  iterative improvement cycle.
  }
  \label{fig:architecture}
\end{figure}

In this section, we present our Recon-Act multi-agent pipeline 
both during training and inference time, as depicted in \cref{fig:architecture}.

Our pipeline comprises two integrated teams, namely
Reconnaissance Team and Action Team. 
In accordance with its role specification, 
the Reconnaissance Team gives requirements and advices
based on its intelligence gatherd through reconnaissance.
These intelligence include failed trajectories derived from the 
Action Team’s interactions with the environment,
successful trajectories from training set, 
and browser contexts along the trajectories.
The Reconnaissance Team identifies the causes of failure, 
and create or update a tool that is helpful to the Action Team in solving the task. 
Once such tool is registered, 
the Action Team immediately receives their specifications, 
thereby acquiring the capability to invoke them in real time.

During training stage,
The Reconnaissance Team analyzes the problem 
and incrementally augments and updates the toolset, 
thereby progressively enhancing the system’s 
cross-task generalization and decision-making capabilities. 
When tool addition or update happens, 
the system performs an inference pass on the 
Training Set to obtain additional Trajectories. 
The training process iterates until 
the Reconnaissance Team can no longer augment or update the toolset, 
or until no improvement in Training Set accuracy is observed after several times of consecutive tool updates, 
at which point training is terminated.

During inference stage, 
only the Action Team performs the task,
which can invoke pretrained (automatically generated) tools
to address typical issues encountered during task execution. 
It fully leverages the available actions to increase 
success rates while substantially improving runtime efficiency.

\subsection{Reconnaissance Team}
\label{subsec:reconnaissance_team}

For Reconnaissance Team, 
we predefine a set of cold-start queries derived from 
real users' needs in browser-use scenarios
which contains both successful and failed trajectories for
one query.
They target common websites and exhibit a degree of generalizability. 
Crucially, the queries used to train the 
Reconnaissance Team are problem instances that 
the system cannot currently solve.
This is necessary to establish a feedback 
loop along with the successful trajectories, 
without which learning cannot proceed. 

We design the reconnaissance agent as a lightweight multi-agent system
comprising two components, 
an Analyst and a Coder, 
alongside a built-in reconnaissance toolkit.
The toolkit includes basic web observation tools such as 
get url, image, and SOM(set-of-marks) observations (in text) etc., 
which enable page-structure parsing, 
extraction of visual cues, 
and related capabilities.
The Analyst performs a degree of information compression: 
conditioned on the task specification and the error categories provided by an evaluator, 
it compares erroneous and successful trajectories at the step level, 
selects and invokes appropriate reconnaissance tools, 
infers the root causes of failures, and proposes remedial strategies. 
The Coder then maps these requirements and 
operational procedures into executable code. 
In conjunction with a tool registration mechanism, 
all tools conform to a unified parameter schema and output format: 
they accept a superset of potentially relevant arguments and return a string. 
This design avoids per-task parameter customization and thereby reduces coding complexity.

The Reconnaissance Team is active only during training, 
which proceeds in iterative Rollout, Evaluate, Generate, Update cycles. 
First, we execute cold-start queries to get new failure trajectories. 
These failure trajectories, together with available success trajectories 
and the current browser context, 
are provided to the Analyst. 
Through contrastive, step-level analysis, the Analyst synthesizes tool specifications. 
The Coder then generate the tool and submits a registration request. 
Once the Tool Manager completes registration, 
the Action Team performs a full inference episode. 
During inference, each action produced by the Action Team is executed 
against the browser via the Playwright API. 
After the episode concludes, if the resulting trajectory is judged 
correct by the evaluator, training for that instance terminates.

\subsection{Action Team}
\label{subsec:action_team}

The Action Team comprises three components: 
a Master, a Tool Manager, and an Execution Agent. 
The Master interprets the user query and 
the browser context to identify the current subtask 
and determines whether to invoke a tool, as well as which tool to invoke. 
The Tool Manager functions essentially as a coding agent. 
When the Reconnaissance Team issues a tool registration request, 
the Tool Manager decides, based on the full set 
of conditions and the tool’s implementation,
whether to add a new tool or update an existing one. 
Updated tools incorporate conditional logic 
to avoid altering the behavior of prior tool invocations. 
These updates are active only during the training stage 
and are disabled at inference time. 
During inference, to ensure the effectiveness of the tool and the scenario generalization 
capability of the entire system,
we added a hard-coded tool-routing mechanism.
The Execution Agent serves as a comprehensive fallback: 
it can generate one of the actions in the entire action space 
and thus guarantees a default output. 
If a tool call fails or no tool is invoked, 
the final action is taken from the Execution Agent’s output. 
Tools can be registered in two modes: 
Hint and Decision. 
A Hint-mode tool which is less deterministic or more context-sensitive, 
returns reconnaissance signals to the 
Execution Agent to improve task completion, 
whereas a Decision-mode tool with consistently stable behavior,
directly emits an action from the action space. 
Outputs from decision-mode are authoritative.
Whenever a Decision tool produces an action, 
the system executes it as specified.

At the start of an inference episode, 
upon receiving the initial query and browser context, 
the Master first interprets the query and context and selects a tool to invoke. 
The tool router then dispatches and executes the corresponding tool. 
If the routed tool is in Hint mode, 
the system executes the Execution Agent afterward to obtain the final action. 
If the routed tool is in Decision mode, its action is returned directly. 
The emitted action interacts with the browser environment to update the state, 
yielding a new context for the next step.

\section{Experiments}
\label{sec:experiments}

In this section, we conduct some experiments to evaluate the performance of our proposed Recon-Act.

\begin{table}[htbp]
\centering
\caption{Success rates of baseline LLM and VLM agents on VisualWebArena}
\label{table:main_results}
\resizebox{\textwidth}{!}{
\begin{tabular}{l l l c c c c}
\toprule
\multirow{2}{*}{\textbf{Paper}} &
\multirow{2}{*}{\textbf{Method}} & 
\multirow{2}{*}{\textbf{Model}} & 
\multicolumn{4}{c}{\textbf{Success Rate ($\uparrow$)} (\%)} \\
\cmidrule(lr){4-7}
& & & Classifieds & Reddit & Shopping & Overall \\
\midrule
$\text{VWA}^1$ & Multimodel (SoM) Image + Caps + SoM & Gemini-Pro & 3.42 & 3.81 & 7.73 & 5.71 \\
$\text{VWA}^1$ & Multimodel Image + Caps + Acc. Tree & Gemini-Pro & 3.42 & 4.29 & 8.15 & 6.04 \\
$\text{VWA}^1$ & Text-only Acc. Tree & GPT-4 & 5.56 & 4.76 & 9.23 & 7.25 \\
$\text{VWA}^1$ & Caption-augmented Acc. Tree + Caps & GPT-4 + BLIP-2-T5XL & 8.55 & 8.57 & 16.74 & 12.75 \\
$\text{VWA}^1$ & Multimodel Image + Caps + Acc. Tree & GPT-4V & 8.12 & 12.38 & 19.74 & 15.05 \\
$\text{VWA}^1$ & Multimodel (SoM) Image + Caps + SoM & GPT-4V & 9.83 & 17.14 & 19.31 & 16.37 \\
$\text{WebDreamer}^2$ & - & Qwen2-VL-7B & 17.9 & 11.1 & 20.2 & 17.20 \\
$\text{WebDreamer}^2$ \quad & - & Qwen2-VL-72B & 19.6 & 15.9 & 24.6 & 21.00 \\
$\text{WebDreamer}^2$ \quad & - & Dreamer-7B & 21.4 & 15.9 & 25.4 & 21.90 \\
$\text{ICAL}^3$ & - & GPT-4V & - & - & - & 22.70 \\
$\text{WebDreamer}^2$ \quad & - & Dreamer-7B + In-Domain & 25.0 & 15.9 & 26.3 & 23.20 \\
$\text{WebDreamer}^2$ \quad & - & GPT-4o & 23.2 & 17.5 & 26.3 & 23.20 \\
$\text{ICAL}^3$ & - & GPT-4o & - & - & - & 23.40 \\
$\text{TreeSearch}^4$ & Search + SoM & GPT-4o & 26.5 & 20.5 & 29.0 & 26.40 \\
$\text{ExAct}^5$ & MCTS SA SoM + Caption + Image & GPT-4o & 37.6 & 23.8 & 29.4  & 30.22 \\
$\text{ExAct}^5$ & R-MCTS SA SoM + Caption + Image & GPT-4o & 40.2 & 25.2 & 31.9 & 32.53 \\
$\text{ExAct}^5$ & R-MCTS MAD SoM + Caption + Image & GPT-4o & \textbf{41.0} & \textbf{28.7} & 32.3 & 33.74 \\
\midrule
ours & Recon-Act & GPT-5-Chat & 39.32 & 27.14 & \textbf{39.27} & \textbf{36.48} \\
\midrule
Human & - & - & 91.07 & 87.10 & 88.39 & 88.70 \\
\bottomrule
\end{tabular}
}
\begin{tablenotes}[flushleft]
\footnotesize
\item[]1 \citet{koh2024visualwebarenaevaluatingmultimodalagents}.
2 \citet{gu2025llmsecretlyworldmodel}.
3 \citet{sarch2025vlmagentsgeneratememories}.
4 \citet{koh2024treesearchlanguagemodel}.
5 \citet{yu2025exactteachingaiagents}.
\end{tablenotes}
\end{table}

\subsection{Datasets and Evaluation Metrics}
\label{subsec:datasets}

We evaluated our method on the VisualWebArena
\citep{koh2024visualwebarenaevaluatingmultimodalagents} 
dataset,
which is a benchmark for evaluating agents that can understand and act 
upon the visual content of the web. 
It targets realistic tasks 
requiring joint reasoning over text and images, 
such as selecting a reasonably priced used car on classifieds 
or comparing sellers and lowest prices 
for a specified product across websites. 
The dataset comprises approximately 910 queries spanning three domains:
classifieds, shopping, and reddit forum. 
Its evaluation supports multiple criteria: 
exact match (predictions must exactly match the reference), 
must-include (predicted key points must be covered by the reference), 
semantic equivalence (judged by a large language model), 
prohibited-content checks (any forbidden item in the output constitutes failure)
and visual question answering–style assessment to determine goal completion, 
complemented by fuzzy image matching using the 
Structural Similarity Index (SSIM) to 
evaluate the closeness of captured or localized images.

\subsection{Main Results}
\label{subsec:main_results}


The results on VisualWebArena dataset are presented in 
\cref{table:main_results}.
We achieves an overall success rate of 36.48\%, 
surpassing the previous best 
\citep{yu2025exactteachingaiagents} by 2.74\%.
For subdomain, we obtain 39.27\% on Shopping, 
substantially outperforming the prior best of 32.30\% (+6.97\%).
On Classifieds and Reddit, 
we trailing the current baselines (41.00\% and 28.70\%) 
by only 1.68\% and 1.56\% respectively.
Compared with earlier methods such as 
\citet{koh2024treesearchlanguagemodel,gu2025llmsecretlyworldmodel,sarch2025vlmagentsgeneratememories}, 
our overall improvements are typically above 10\%.
While a gap to human performance remains, 
these results set Recon‑Act as the 
new state of the art on this benchmark.

\subsection{Implementation Details}
\label{subsec:implementation_details}

Our system does not incorporate random-walk-based autonomous 
exploration as in \citet{gu2025llmsecretlyworldmodel}. 
Instead, guided by the coverage of our target datasets, 
we manually authored a small training set, 
with fewer than 10 examples per domain. 
We argue that random-walk exploration tends to 
produce overly large corpora with substantial redundancy, 
which is misaligned with our efficiency and curation goals.
Based on the training data and Level 3 configuration, 
we implemented a total of 5 agents (as 2 teams) and 
11 tools, as summarized in \cref{table:agents} and \cref{table:tools}.
Among the agents, 
the coder, master, and execution agent 
are driven by large models, 
while the analyst and tool manager are driven by human.

\begin{table}[h]
\centering
\caption{Agents in Recon-Act pipeline}
\label{table:agents}
\footnotesize
\begin{tabularx}{\linewidth}{l l l X}
\toprule
\textbf{Team} & \textbf{Agent Name} & \textbf{Driven By} & \textbf{Functionality} \\
\midrule
\multirow{2}{*}{Reconnaissance}
& Analyst & Human & 
Compares trajectories at the step level, 
calls appropriate reconnaissance tools, 
infers the cause of failure and 
proposes remediation strategies \\
& Coder & GPT-5-Chat & 
Transfer remediation strategies into tool codes \\
\midrule
\multirow{3}{*}{Action}
& Master & GPT-5-Chat & 
Interprets the query and context and selects a tool to invoke \\
& Tool Manager & Human & 
Decides whether to add a new tool or update an existing one
and updates tools with conditional logic \\
& Execution Agent & GPT-5-Chat & 
Generates a default action  \\
\bottomrule
\end{tabularx}
\end{table}



\begin{table}[htbp]
\centering
\caption{Tools created by Recon-Act}
\label{table:tools}
\footnotesize
\begin{tabularx}{\linewidth}{l l X}
\toprule
\textbf{Tool Name*} & \textbf{Type} & \textbf{Functionality (Description)} \\
\midrule
AuthorFinder & Decision & who can find the author's all post when you are at the post detailed page or commemt page, and go to the author's page \\
CategoryGuide & Decision & who can guide you to the specific category page when you are at the shopping site, can only be called on shopping site \\
ClassifiedsPriceSorter & Decision & who can sort items in classifieds site according to intent, can only be called on classifieds site, should always be called after every action \\
DownVoter & Decision & who would downvote the current post when you are at the post detailed page or comment page \\
ImageSearcher  & Decision & who can find the most similar post to the input image on reddit and go to its detailed page and commemt page, should only be called on reddit site \\
ShoppingImageFinder & Decision & who can find the desired image and go to its detailed page \\
ShoppingPriceSorter & Decision & who can sort items in shopping site according to intent, can only be called on product list pages \\
SubRedditNavigator & Decision & who can navigate to the subreddit page when you are at the post detailed page or comment page \\
UpVoter & Decision & who would upvote the current post when you are at the post detailed page or comment page \\
PostTimeFinder & Hint & who can find the post time when you are at the post detailed page or comment page \\
RedditImageDescriptor & Hint & who can return you the image description of the post image when you are at the post detailed page or comment page, can only be called on reddit site \\
\bottomrule
\end{tabularx}
\begin{tablenotes}[flushleft]
\footnotesize
\item[]* The tool names may be a bit chaos, but we have kept all the original names and the descriptions are exactly the same as what we gave to the master.
\end{tablenotes}
\end{table}





The master and executor operate purely based on prompts.
For the coder in Reconnaissance Team, 
we fixed the input–output interface and the basic code structure
to ensure that tool codes can be generated with high feasibility.

For analyst in reconnaissance team, 
we need to prompt the model to ground its reasoning 
on concrete solution procedures.
We prefer to direct the model to navigate straight to 
the target action (e.g. the ``goto'' operation) 
rather than relying 
on click-based exploration, which is less dependable. 
For the same reason, we also prompt to design sorters and image searcher tool 
to operate via ``goto'' action when possible.

For tool manager in action team, 
the code merge happens in registration is the toughest work.
Automatically generated tools are frequently narrow and fragmented. 
For example, 
a price sorter created for a site that only requires 
finding the cheapest item will contain only a ``cheapest'' branch, 
while a subsequent query requesting the most expensive item would then fail. 
Without consolidation, 
such specialization leads to tool proliferation. 
Moreover, 
because our agents does not ingest the entire page context at once, 
the scope of a tool can be ambiguously defined: 
a tool named ``price sorter'' might only support low-to-high sorting; 
a Reddit voting tool with only down voting function
might be labeled simply as ``voter''.
Because of the above insights, 
humans are currently involved in naming tools, 
adjusting feature branches, 
and merging tools where appropriate.

It's important to note that
because each website has its own unique characteristics, 
we specifically have the reconnaissance team pay attention 
to the specific website when writing tools,
which ensures that there is no confusion when calling similar tools.

\section{Conclusion}
\label{sec:conclusion}

We introduce Recon-Act, a tool-centric, self-evolving system 
for browser interaction that relies on a 
dual-team reconnaissance–action collaboration. 
We formalize in-browser reconnaissance, 
enabling the agents to distill salient intelligence 
from information-dense pages with a small number of 
exploratory actions and to generate feedback through 
contrastive analysis of positive and negative instances, 
thereby establishing a closed-loop evolutionary pipeline 
spanning data, tools, actions, and feedback. 
The system adopts a staged experimental paradigm to 
progressively realize capabilities and currently reaches 
Level 3: human–AI collaboration is retained for analyst and tool manager, 
while the remaining components are driven by 
vision–language models.
Under this configuration, 
Recon-Act sets a new state of the art on VisualWebArena, 
demonstrating effectiveness and efficiency in autonomously 
acquiring cues, invoking tools, 
and completing complex multi-turn tasks in unfamiliar environments.

\section{Limitations and Future Work}
\label{sec:limitations_and_future_work}

To realize intelligence beyond Level 5, 
our future work will proceed along the following directions:
\begin{enumerate}[leftmargin=12pt, nosep]
  \item \textbf{Increasing Autonomy: }
Present learning capability is heavily dependent on 
our constructed training data, 
particularly on the inclusion of successful trajectories,
making this training process similar to ``supervised'' training.
We plan to prompt the model to conduct random-walk-style 
self-exploration in order to generate additional successful trajectories. 
This in turn, will make the construction of the training set more autonomous.
  \item \textbf{Reasoning and Coding Skills: }
To progress from Level 3 to higher levels and reduce reliance on 
human analysis and tool management, 
we must further strengthen the analyst and tool manager components. 
The analyst encapsulates reasoning ability, 
while the tool manager reflects coding competence. 
For the analyst, 
we intend to collect a targeted corpus of analytical data 
and train it with diverse browser contexts so that it acquires robust, 
context-aware analytical skills. 
It should not only give insights related to the task, 
but also consider steps to reduce the difficulty 
of the task for itself, 
so as to make the subtask more suitable for large models.
An example arises on Classifieds websites, 
where image-localization steps 
consistently select the wrong bounding box
because of the small size of the image. 
Guided by cold-start trajectories, 
we switches the presentation from list view to grid view.
The grid layout enlarges thumbnails, 
thereby reducing the difficulty for a VLM to interpret the images. 
Moreover, given the URLs before and after the layout change, 
it should discover that switching from list to grid 
can be achieved not only by clicking a page-level toggle 
but also by appending a fixed-pattern subpath
to the URL to reach the corresponding layout page directly. 
This can be concluded as a tool
and described functionally as enlarging images when 
they are otherwise too small.
For the tool manager, the bottleneck lies in 
the complexity of branching and iterative 
code modification during the registration workflow, 
specifically maintaining isolation between existing capabilities and 
newly introduced ones via feature branches. 
Additionally, 
the master agent still has a certain probability of error 
when calling the tool.
If we consider making the subtask suitable for large models again,
reducing the number of tools that need to be called 
by merging the functions of the tools 
can make the orchestration easier 
without having to increase the master's orchestration ability.
We will address this through 
similarly targeted training in the future.
  \item \textbf{Expanding Reconnaissance Capabilities: }
Our current reconnaissance module performs successfully 
only on a fixed set of websites and 
has not yet generalized to a broader, more heterogeneous web environment. 
\end{enumerate}


\newpage
\bibliographystyle{config/antgroup}
\bibliography{references}


\end{document}